# A High-Performance Adaptive Quantization Approach for Edge CNN Applications

Hsu-Hsun Chin
Department of Computer Science
National Tsing-Hua University
Hsinchu, Taiwan
kim60138@gmail.com

Ren-Song Tsay
Department of Computer Science
National Tsing-Hua University
Hsinchu, Taiwan
rstsay@gmail.com

Hsin-I Wu
Department of Computer Science
National Tsing-Hua University
Hsinchu, Taiwan
hiwu.dery@gmail.com

## ABSTRACT

Recent convolutional neural network (CNN) development continues to advance the state-of-the-art model accuracy for various applications. However, the enhanced accuracy comes at the cost of substantial memory bandwidth and storage requirements and demanding computational resources. Although in the past the quantization methods have effectively reduced the deployment cost for edge devices, it suffers from significant information loss when processing the biased activations of contemporary CNNs. In this paper, we hence introduce an adaptive high-performance quantization method to resolve the issue of biased activation by dynamically adjusting the scaling and shifting factors based on the task loss. Our proposed method has been extensively evaluated on image classification models (ResNet-18/34/50, MobileNet-V2, EfficientNet-B0) with ImageNet dataset, object detection model (YOLO-V4) with COCO dataset, and language models with PTB dataset. The results show that our 4-bit integer (INT4) quantization models achieve better accuracy than the state-of-the-art 4-bit models, and in some cases, even surpass the golden full-precision models. The final designs have been successfully deployed onto extremely resource-constrained edge devices for many practical applications.

## 1 INTRODUCTION

Deep learning convolutional neural networks (CNNs) have been so successful and are adopted at a breakthrough rate, particularly for computer vision applications, such as image classification [1], object detection [2], and image segmentation [3]. With the popularity of the Internet of Things (IoT), CNNs are highly desirable to almost all IoT application domains, including agriculture [4], flood management [5], smart health monitoring [6]. However, the high-performance state-of-the-art CNNs require a significant amount of parameter volume and computation performance. For instance, the ResNet50 [7] model, commonly used for image classification tasks, has 25.5 million parameters (about 100 MB) and requires 3.9 billion floating-point multiply-and-add (MAC) operations for recognizing an image in the inference step. Unlike the powerful performance of cloud servers, performing inference on low-end System-on-a-Chip (SoC) used in mobile devices or IoT devices is a significant challenge due to the limited computational capability, power consumption, and storage space. In sum, deploying CNNs on resource-constrained platforms is highly demanded but also a considerable challenge.

In order to address this challenge, several approaches have been proposed. Network pruning methods significantly reduce the model size by eliminating redundant and unimportant parameters [8][9][10]. Also, several efficient architectures of CNN have been proposed to balance the accuracy and computation cost [11][12][13][14][15]. However, in the resulting compressed models, only the total number of operations is reduced but not the computational complexity of individual operations. Hence, such models still require heavy computational units, such as floating-point units and loaded hardware resources.

Another popular approach is network quantization, which is the focus of this paper. The quantization method has also been widely adopted to reduce the size and computational complexity of CNN models by converting the weights (parameters) and/or activations (intermediate outputs) from floating-point numbers to low bit-width representations. The low bit-width representation has several advantages in terms of hardware resource requirements. Particularly, the storage requirement is substantially reduced. The memory communication efficiency is also greatly improved due to the use of low bit-width data. In addition, the circuit complexity of the low bit-width MAC (multiply-accumulate) unit is much lower than that of the floating-point MAC unit. Thus the quantization models save both chip area and power consumption significantly. For instance, the floating-point adder consumes 30 times more energy and 116 times more area than the 8-bit adder; the floating-point multiplier consumes 18 times more energy and 27 times more area than the 8-bit multiplier [16].

Despite the high computational efficiency of low bit-width inference, the task accuracy degrades and needs to be recovered by retraining. The experimental results from recent research [17][18] show a comparison of model accuracy under different bit-width quantization in the image classification task, and the results support that the 4-bit quantization models strike the best balance between minimal computing/memory complexity and minimal model accuracy degradation.

Generally, existing quantization approaches achieve good results in the traditional ReLU-based models, where the activations are passed through the ReLU function [19] and then restricted activations to be non-negative values. In contrast, contemporary network architectures adopt more complex activation functions, such as Swish [20], Leaky-ReLU [19], and Mish [21], with which the range of activation distributions is relatively more irregular than the weight distribution and hence the traditional quantization methods experience a significant information loss.

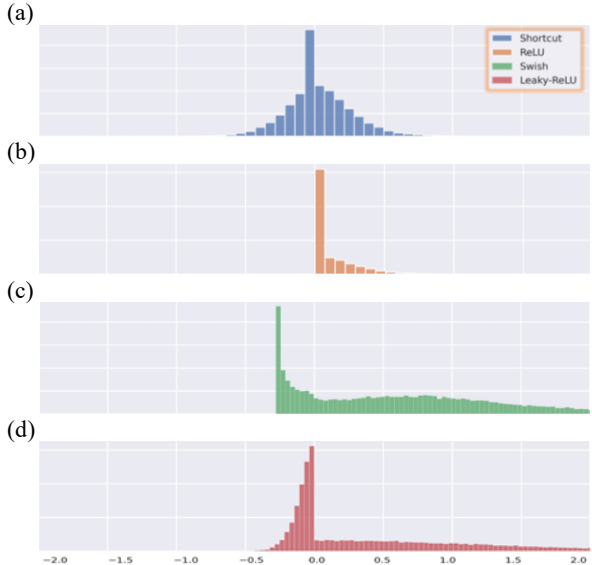

Figure 1: The output distribution of different activation functions. (a) The output of the shortcut layer of ResNet50 model. (b) The output of the ReLU layer of ResNet50 model. (c) The output of the Swish layer of EfficientNet-b0 model. (d) The output of the Leaky-ReLU layer of DarkNet53 model.

Unlike the generally Gaussian-shaped weight distributions, we observe in Fig. 1(b) that the ReLU's outputs are restricted to the non-negative side, i.e., $[0, +\infty)$, while the outputs of the shortcut layer spread through the range $(-\infty, +\infty)$ with an almost symmetric distribution to the value zero, as in Fig. 1(a). In contrast, the Swish's outputs, Fig. 1(c), and Leaky-ReLU's outputs, Fig. 1(d), expand over a wide positive range but a narrow negative range. Therefore, different activation distributions present quite different value ranges and bias values. Practically, the varieties are challenging for low bit-width quantization methods to cover well, whereas 32-bit floating-point representation can accurately express all types of distributions. Nevertheless, the indispensable benefit of reducing memory overhead strongly demands a more robust quantization method to manage the dynamic distribution range.

The traditional uniform quantization methods [17][22][23] simply quantize the normalized distribution using a simple scaling, and the results often suffer significant quantization errors, particularly when processing uneven and biased distributions. For instance, we show in Fig. 2(a) a sample original data distribution with the dashed box denoting the quantization range. If the scaling simply aligns the most positive value to the rightmost boundary of the quantization range with the quantization points marked on the x-axis of the associated histogram as shown in Fig. 2(b). Note that in this case, the four most negative quantization points are wasted because of the biased distribution. On the other hand, if the most negative value is scaled to align the left quantization boundary, then the excessive large positive values beyond the quantization range are truncated with large quantization errors, as indicated by the arrowhead in Fig. 2(c). To alleviate the biased distribution problem, we adopt the unified approach from Jacob et al. [24] by performing both shifting and scaling of the activation distribution before quantization, as illustrated in Fig. 2(d), and achieve a significantly better model accuracy. Notably, thanks to this unified quantization approach, the activations and weights can be represented in the same way, so the hardware structures designed for each layer can be reused, thus saving hardware resources.

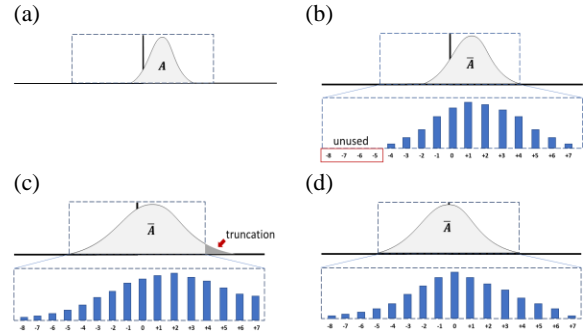

Figure 2: The quantization process of distribution, where the dashed box denotes the quantization range. (a) The original distribution. (b) Quantization after the scaling keeps the entire distribution within the quantization range. (c) Quantization after the scaling makes the distribution fill the entire quantization range. (d) Quantization after scaling and shifting.

Furthermore, improper scaling and shifting could lead to under-utilization of the quantization points or excessive truncations, and quantization errors accumulate over deep layers, ultimately degrade the model accuracy. Consequently, a critical issue is how to find the optimal scaling and shifting. Traditionally, a commonly adopted method is to minimize the quantization error following certain metrics, such as signal-to-quantization-noise ratio (SQNR) [25], information entropy [26], by the linear search algorithm. Nevertheless, since the linear search algorithm requires traversing all the search space and finding the optimum over a few data batches, the algorithm is not practical for iterative training. In fact, the quantization error metric does not correlate directly to the target task loss, the difference between the prediction and ground truth, a measure of the model accuracy. Therefore, inspired by the work of PACT [17], we propose a direct training adjustment of the quantization parameters according to the task loss rather than the quantization error metric.

The main contributions of this paper can be summarized as follows:

1. We propose an adaptive uniform quantization method that dynamically performs optimal scaling and shifting before the quantization process to minimize the information loss due to low-bit width quantization and diverse activation distributions.

2. We effectively minimize the quantized information loss by optimized training of the scaling and shifting factors according to the task loss.
3. Our proposed hardware design conveniently reuses the same architecture in each convolutional layer, thus saves significant hardware resources.

The remainder of the paper is organized as follows. Section 2 reviews related works about network compression. Section 3 introduces the basic concept of CNN. In Section 4, we elaborate on our adaptive uniform quantization method. Next, the quantization experiments on several popular CNNs are presented in Section 5. Finally, we conclude in Section 6.

## 2 RELATED WORK

The effectiveness of CNNs depends on the model parameters and operations. Generally, the loaded CNN models critically deter the applications in some resource-constrained scenarios. For instance, embedded systems with limited resources have to implement downsized CNN models with reduced model accuracy. Therefore, extensive research works have been conducted on model compression to minimize the overhead of inferencing with good accuracy. We now briefly review three types of popular approaches, architecture optimization, pruning, and quantization.

### 2.1 Architecture Optimization Methods

This type of approach focuses on optimized architecture designs. For real-time applications, the depth-wise separable convolution and inverted residual structure are widely adopted in CNNs. Examples such as MobileNet [14] have effectively improved the accuracy at a lower computation cost. Another popular approach for neural network designs is network architecture search [27][28][29], i.e., a representative framework AutoML (automated machine learning) [15], which automatically searches for the best-performing architecture considering the computation cost. Under the same accuracy measure, the aforementioned network architectures generally require fewer operations and parameters than traditional network architectures. However, the complexity of each operation is not affected and reduced. It still requires floating-point arithmetic to execute such a model and aggravates the overhead of hardware resources. Generally, it is a trend to combine optimized architectures and quantization approaches, replacing floating-point arithmetic with integer arithmetic [30].

### 2.2 Network Pruning Methods

The pruning methods attempt to eliminate unimportant weights in a pre-trained model while maintaining model accuracy after retraining. These approaches apply various weight importance measures for pruning evaluations. For instance, the earliest network pruning method [9][31] prunes the network parameters based on the Hessian loss function. Deep Compression [8] simply removes small weights, as they observed that small weights have little impact on model accuracy. Most other pruning approaches adopt L1 or L2-norm to identify unimportant weights [32-34]. Although the resulting sparse weight matrices are mostly zeros, to store and manage non-zero values, a dedicated data structure and a complicated sparse matrix multiplication method are required, and the sparse matrix methods cannot be accelerated without specific hardware designs [35]. Therefore, the induced design overheads from network pruning have greatly hindered the feasibility for applications on embedded devices with limited resources.

### 2.3 Network Quantization Methods

The quantization approaches attempt to use a reduced number of data bits to represent the models. The quantization approaches have been actively studied as they generally can achieve feasible results. Works, such as BinaryConnect [36], Binary-Weight-Network [37], TWN [38], TTQ [39], significantly reduced the model size using only one or two bits for the model *weight* bit-width. Unfortunately, with too few bits for model representation, these works suffered an irreparable model accuracy loss.

In contrast, several works aim to maintain high model accuracy by considering both weight and activation quantization with low bit-width. Jacob and Krishnamoorthi [24][40] demonstrated that quantization using 8-bit integer representations is good for most networks to maintain model accuracy with reduced resources. Hwang and Lin [25][41] adopted a fixed-point CNN model representation and significantly reduced memory and hardware resource usage. However, the low bit-width fixed-point format is prone to overflow or underflow issues. To resolve this issue, a few works [42] applied a scaling factor in each layer to cover different distribution ranges. Nevertheless, the training process may shift the parameter distribution range and degrade the effectiveness of the fixed scaling factors. Therefore, our proposed approach also trains the scaling factors and optimizes them for changing parameter ranges.

Recently, extensive research works have focused on activation quantization in addition to *weight* quantization since the *activations* also take memory space. Notably, PACT proposed a parameterized and learnable clipping activation function to balance the truncation and quantization errors, with the quantization interval determined by the clipping parameter. Based on PACT, NICE [22] introduced noise to emulate the quantization effect and set the initial clamping parameter of activation through statistics to prevent the network from falling into a local minimum. In contrast, LSQ [23] and QIL [43] proposed to select the quantization intervals for both *weights* and *activations* according to the difference between the prediction and the ground truth.

However, the above-mentioned quantization methods consider only ReLU-based activation models and could not effectively handle the latest developed CNN models with more complicated activation functions, which produce very different output distributions, as shown in Fig. 1. Note that ReLU limits the activation values in the interval $[0, +\infty)$. Our proposed quantization approach can robustly manage different activation functions with any distributed range.

## 3 PRELIMINARY

Before we explain our approach, we first define the notations used in this paper and review the working principle of convolutional neural networks.

In general, convolutional neural networks perform a series of matrix multiplication in each layer. Given an L-layers convolutional neural network, for each layer $l$, $0 \leq l < L$, we have

$$\mathbf{Y}^l = \mathbf{W}^l * \mathbf{A}^l + \mathbf{b}^l \quad (1)$$

$$\mathbf{A}^{l+1} = g^l(\mathbf{Y}^l) \quad (2)$$

where $\mathbf{W}^l$ denotes the *weight* matrix, $\mathbf{A}^l$ denotes the input activation matrix, and $\mathbf{b}^l$ denotes the bias vector. Note that $g^l(.)$ is a non-linear activation function (e.g., ReLU, Leaky-ReLU) and the intermediate result $\mathbf{Y}^l$ is passed through the non-linear activation function $g^l(.)$ to obtain the activation matrix $\mathbf{A}^{l+1}$ for the next layer. For simplicity, we assume that all matrices are 2-dimensional N×N and the matrix element of the *l*-th layer is indexed as $\mathbf{W}_{ij}^l$ for $0 \leq i, j < N$.

## 4 THE PROPOSED METHOD

The main objective of our proposed method is to design a unified quantization method to cover the parameter range of each layer dynamically and use low bit-width integer arithmetic to execute the operations throughout all inference layers while maintaining model accuracy. Furthermore, our method is a two-stage neural network training process: a forward propagation stage and a backward propagation stage. The forward propagation stage follows the inference operations of the quantized model and obtains the task loss for optimization in the backward propagation stage, while the task loss is used for adjustment of the quantization parameters. In this way, we dynamically update the quantization parameters related to scaling and shifting factors based on the task loss.

### 4.1 Forward Propagation

With the uniform quantization, the input, output, and *weight* values of each layer are quantized into 4-bit integers in $[-8, +7]$ so that we can have an effective unified hardware component for CNN inference computations in each layer. Nevertheless, to utilize the available quantization range fully, proper scaling, or normalization, is critical.

For example, in order to quantize the *weight* of l-th layer using the 4-bit integer format, whose range is $[-8, +7]$, we first calculate the normalized *weight* matrix $\overline{\mathbf{W}}^l$ as the following,

$$\overline{\mathbf{W}}^l = \frac{\mathbf{W}^l}{f_W^l} \Rightarrow \mathbf{W}^l = f_W^l \overline{\mathbf{W}}^l \quad (3)$$

where $f_W^l$ denotes the scaling factor of *weight* and the optimal scaling factor $f_W^l$ is adjusted according to the gradient calculation in the backward propagation, which will be described later. Then the quantized matrix is obtained from the normalized matrix as the following,

$$\widehat{\mathbf{W}}^l = q_{int4}(\overline{\mathbf{W}}^l) \quad (4)$$

where the quantization function $q_{int4}(.)$ is defined as the following,

$$\hat{x} = q_{int4}(x) = \begin{cases} +7, & x > 7 \\ -8, & x \leq -8 \\ \text{Round}(x), & \text{otherwise} \end{cases} \quad (5)$$

where the function Round(.) rounds a number to its nearest integer.

In addition to *weight* values, activation values also take considerable memory space and need to be quantized [25]. Unlike the *weight* distribution, which is usually bell-shaped and concentrates around zero value, the activation distribution of each layer is different, depending on the activation function $g^l$ as illustrated in Fig. 1. Therefore, when quantizing the activation matrix into the 4-bit integer format, to cover distinct distribution ranges, the activation matrix $\mathbf{A}^l$ is first shifted and then scaled to obtain the normalized activation matrix $\overline{\mathbf{A}}^l$ as:

$$\overline{\mathbf{A}}^l = \frac{\mathbf{A}^l - z_A^l}{f_A^l} \Rightarrow \mathbf{A}^l = f_A^l \overline{\mathbf{A}}^l + z_A^l \quad (6)$$

where $f_A^l$ denotes the scaling factor of activation and $z_A^l$ denotes the zero-point of activation and is used to eliminate the bias by shifting. We will discuss the process of finding the optimal scaling factor $f_A^l$ and zero-point $z_A^l$ along with $f_A^l$ in the backward propagation stage.

$$\widehat{\mathbf{A}}^l \equiv q_{int4}(\overline{\mathbf{A}}^l) \quad (7)$$

Although for the l-th layer, both the quantized input activation matrix $\widehat{\mathbf{A}}^l$ and weight matrix $\widehat{\mathbf{W}}^l$ are in 4-bit integer format, after the layer convolution computations, the output activation may overflow the 4-bit integer range. Therefore, the output activation $\mathbf{A}^{l+1}$ needs to be re-normalized and shifted as the following,

$$\mathbf{A}^{l+1} = f_A^{l+1} \overline{\mathbf{A}}^{l+1} + z_A^{l+1} \quad (8)$$

Next, by combing Eq.(2) and Eq.(8), we have

$$g^l(\mathbf{W}^l * \mathbf{A}^l + \mathbf{b}^l) = f_A^{l+1} \overline{\mathbf{A}}^{l+1} + z_A^{l+1} \quad (9)$$

Then, based on Eq.(3) and Eq.(6), the left side of Eq.(9) becomes

$$g^l(f_W^l \overline{\mathbf{W}}^l * (f_A^l \overline{\mathbf{A}}^l + z_A^l) + \mathbf{b}^l)$$

$$= g^l(f_W^l \overline{\mathbf{W}}^l * (f_A^l \overline{\mathbf{A}}^l + z_A^l) + \mathbf{b}^l)$$

$$= g^l(f_W^l f_A^l \overline{\mathbf{W}}^l * \overline{\mathbf{A}}^l + \tilde{\mathbf{b}}^l) \quad (10)$$

where the new bias $\tilde{b}^l$ denotes $b^l + f_W^l z_A^l \overline{W}^l$.

Finally, from Eq.(9) and Eq.(10), we have the normalized output activation matrix of l-th layer as the following,

$$\overline{A}^{l+1} = \frac{g^l(f_W^l f_A^l \overline{W}^l * \overline{A}^l + \tilde{b}^l) - z_A^{l+1}}{f_A^{l+1}} \quad (11)$$

Then we obtain the quantized output activation $\widehat{A}^{l+1}$ of l-th by normalizing and then quantizing to a 4-bit integer format as:

$$\widehat{A}^{l+1} = q_{int4}\left(\frac{g^l(f_W^l f_A^l \widehat{W}^l * \widehat{A}^l + \tilde{b}^l) - z_A^{l+1}}{f_A^{l+1}}\right) \quad (12)$$

As shown in Fig. 3, we have a complete execution flow of each layer. In general, an intermediate output through convolution and activation function $g^l$ is normalized by shifting and scaling and then re-quantized into the 4-bit integer format before passing as the input to the next layer. Note that the values of $f_A^{l+1}$, $f_W^l$, $f_A^l$, $\tilde{b}^l$ and $z_A^{l+1}$ are all in 16-bit fixed-point format for proper accuracy, while these parameters take only less than 0.1% of the total memory space.

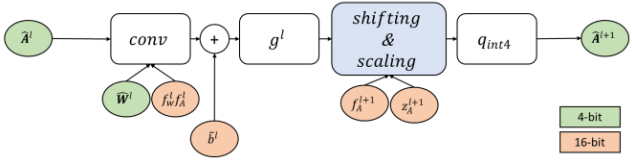

**Figure 3: An overview of integer arithmetic inference of each layer in CNN. The hardware design reuses the same architecture in each convolutional layer, thus saves significant hardware resources.**

With the above computing procedure, the entire CNN inferencing process can be efficiently executed in integer arithmetic. The experimental results support that the 4-bit quantization strikes the best balance between minimal computing/memory complexity and minimal model accuracy degradation. Next, we discuss how to find the optimal quantization parameters.

### 4.2 Backward Propagation

We leverage the backward propagation stage and update the quantization parameters, i.e., scaling and shifting parameters, along with weight values for minimal task loss. The direction and number of updates are determined by the gradient calculated layer-by-layer based on the chain rule. Given the task loss $\mathcal{L}(P, G)$, the gap between the prediction result $P$ and the ground-truth $G$, the gradient of the $l$-th quantization parameter $q^l$ with respect to the task loss is derived based on the chain rule as the following,

$$\frac{\partial \mathcal{L}(P,G)}{\partial q^l} = \frac{\partial \mathcal{L}(P,G)}{\partial A^{L-1}} \cdots \frac{\partial A^{l+1}}{\partial Y^l} \frac{\partial Y^l}{\partial q^l} \quad (13)$$

Then, the $l$-th quantization parameter $q^l$ is updated as:

$$(q^l)_t = (q^l)_{t-1} - \eta * \frac{\partial \mathcal{L}(P,G)}{\partial (q^l)_{t-1}} \quad (14)$$

where $\eta$ denotes the learning rate and $t$ denotes the current iteration of learning.

However, since the quantization process is non-differentiable, the gradient is zero almost everywhere. Thus, we adopt the STE method [44] to estimate the gradient of the quantization function $q_{int4}(.)$ as:

$$\frac{\partial q_{int4}(x)}{\partial x} = \begin{cases} 1, & -8 < x < +7 \\ 0, & otherwise \end{cases} \quad (15)$$

Note that, when using the chain rule for series of operations with the quantization function $q_{int4}(.)$, the gradient of $q_{int4}$ is passed to the next operation only if the input is within the range $[-8, +7]$. The gradient calculation of the quantized weight values is performed in the same way as that of the quantized activations except with the zero-point fixed at zero or no shifting.

We now discuss how to adaptively optimize the scaling factors and zero-points according to the gradients computed from the task loss. Given the task loss $\mathcal{L}(P_q, G)$, i.e., the gap between the prediction result $P_q$ from the quantized network and the ground-truth $G$, we approximate the $l$-th layer activation $A^l$ as

$$A^l = f_A^l \overline{A}^l + z_A^l \approx f_A^l \widehat{A}^l + z_A^l \quad (16)$$

Then, the partial derivatives of the activation matrix $A^l$ with respect to the scaling factor $f_A^l$ and zero-point $z_A^l$ are as the following.

$$\frac{\partial A^l}{\partial f_A^l} \approx \widehat{A}^l + f_A^l \frac{\partial \widehat{A}^l}{\partial f_A^l} \quad (17)$$

$$\frac{\partial A^l}{\partial z_A^l} \approx f_A^l \frac{\partial \widehat{A}^l}{\partial z_A^l} + 1 \quad (18)$$

Since $\widehat{A}^l \equiv q_{int4}(\overline{A}^l)$, following Eq. (15), we have

$$\frac{\partial \widehat{A}^l}{\partial f_A^l} = \frac{\partial \widehat{A}^l}{\partial \overline{A}^l} \frac{\partial \overline{A}^l}{\partial f_A^l} = -(A^l - z_A^l)/f_A^{l^2} \cdot \begin{cases} 1, & -8 < \overline{A}^l < +7 \\ 0, & otherwise \end{cases}$$

$$= -\overline{A}^l/f_A^l \cdot \begin{cases} 1, & -8 < \overline{A}^l < +7 \\ 0, & otherwise \end{cases} \quad (19)$$

$$\frac{\partial \widehat{A}^l}{\partial z_A^l} = \frac{\partial \widehat{A}^l}{\partial \overline{A}^l} \frac{\partial \overline{A}^l}{\partial z_A^l} = \begin{cases} -1/f_A^l, & -8 < \overline{A}^l < +7 \\ 0, & otherwise \end{cases} \quad (20)$$

Finally, combining Eq. (17) and Eq. (19), we have the gradient of $\mathcal{L}(P_q, G)$ with respect to the scaling factor $f_A^l$ as

$$\frac{\partial \mathcal{L}(P_q,G)}{\partial f_A^l} = \frac{\partial \mathcal{L}(P_q,G)}{\partial A^l} \frac{\partial A^l}{\partial f_A^l}$$

$$= \frac{\partial \mathcal{L}(P_q,G)}{\partial A^l} \cdot \begin{cases} \widehat{A}^l - \overline{A}^l, & -8 < \overline{A}^l < +7 \\ \widehat{A}^l, & otherwise \end{cases} \quad (21)$$

Similarly, based on Eq. (18) and Eq. (20), we have the gradient of $\mathcal{L}(P_q, G)$ with respect to the zero-point $z_A^l$ as

$$\frac{\partial \mathcal{L}(P_q,G)}{\partial z_A^l} = \frac{\partial \mathcal{L}(P_q,G)}{\partial A^l} \frac{\partial A^l}{\partial z_A^l}$$

$$= \frac{\partial \mathcal{L}(P_q,G)}{\partial A^l} \cdot \begin{cases} 0, & -8 < \overline{A}^l < +7 \\ 1, & otherwise \end{cases} \quad (22)$$

We observe from Eq. (21) and Eq. (22) that the adjustment process considers not only the task loss but the quantization error. The term $\widehat{A}^l - \overline{A}^l$ from Eq. (21) helps to draw closer the scaled values to the quantization points. On the other hand, the zero-point shifts only if there are scaled values placed outside the quantization range.

Through the above calculations, we find the optimal scaling factors and zero-points according to the task loss, which is more direct and effective than the traditional quantization error metric-based approach.

**Table 1: The ImageNet benchmark results are based on top-1, top-5 accuracy measures with the full-precision model as the baseline.**

| Network | Method | Top-1 | Top-5 |
|---|---|---|---|
| ResNet-18 | Baseline | 69.76% | 89.08% |
| | PACT | 69.20% | 89.00% |
| | NICE | 69.80% | 89.21% |
| | QIL | 70.10% | - |
| | DSQ | 69.60% | - |
| | DoReFa | 68.10% | 88.10% |
| | LQ-Net | 69.30% | 88.80% |
| | LSQ | 71.10% | 90.00% |
| | **Ours** | **71.46%** | **90.16%** |
| ResNet-34 | Baseline | 73.30% | 91.42% |
| | NICE | 73.45% | 91.41% |
| | QIL | 73.70% | - |
| | DSQ | 72.80% | 91.30% |
| | LSQ | 74.10% | 91.70% |
| | **Ours** | **75.27%** | **92.31%** |
| ResNet-50 | Baseline | 76.15% | 92.87% |
| | PACT | 76.50% | 93.20% |
| | NICE | 76.50% | |
| | LQ-Net | 75.10% | 92.40% |
| | LSQ | 76.70% | 93.20% |
| | **Ours** | **76.88%** | **93.45%** |
| Mobilenet-V2 | Baseline | 71.88% | 90.29% |
| | PACT | 61.39% | 83.72% |
| | DSQ | 64.90% | - |
| | Ours (without shifting) | 68.76% | - |
| | **Ours** | **69.21%** | **88.96%** |
| EfficientNet-B0 | Baseline | 76.12% | 93.3% |
| | Ours (without shifting) | 71.86% | 90.18% |
| | **Ours** | **73.25%** | **91.84%** |

## 5 EXPERIMENT

To demonstrate the effectiveness of our method, we test on a few representative CNN models, e.g., image classification, object detection, and language model. We use pre-trained full-precision models as the basis in all experiments, convert the activation and weight values to 4-bit integers, and perform retraining. Finally, we compare our results with the full-precision model and other quantization approaches.

### 5.1 Image Classification

We first apply our adaptive method to popular models, such as ResNet-18/34/50, MobileNet-V2, EfficientNet-B0. For test data, we adopt ImageNet [45], a large-scale image database with thousands of object categories and millions of images, which is has been widely used for visual classification task benchmarks.

Then we benchmark our results with the full-precision network as the baseline and other representative quantization methods, such as PACT, NICE, and LSQ. We list the summary in Table 1. All comparisons are based on Top-1/Top-5, which are the accuracy measures of the ground truth of Top-1/Top-5 predictions.

For all classification experiments, we retrain 90 epochs with an initial learning rate of $10^{-3}$, weight decay of $10^{-4}$ and SGD [46] optimizer. Note that the first and last layer is quantized to 8-bit while all other layers are in 4-bit for a fair comparison.

In summary, our method consistently outperforms all other quantization methods, including the LSQ uniform quantization method [23], which normalizes the distributions only by scaling and is the nearest contender.

Additionally, the results based on our method often surprisingly are better than the full-precision models (the baseline). We believe the reason why our approach performs so well is that our method eliminates the model redundancy and avoids overfitting while significantly enhances the model regularization capability following Ockham's razor principle [47] with no unnecessary assumptions.

Note that we particularly test MobileNet and EfficientNet, two famous slim models whose parameter sets are extremely compact, in order to understand the effectiveness of our method on slim models. The benchmarks show that the traditional quantization methods fail to maintain good model accuracy because that the parameters of slim models have minimal redundancy and are very sensitive to quantization. Nevertheless, our method still performs well on slim models. In general, the Top-1 accuracy rate of our approach is 0.45% to 7.82% better than other approaches. Interestingly, for our method, normalization with shifting and scaling achieves 0.45% to 1.39% better results than that with only scaling.

### 5.2 Object Detection

For the object detection model, we choose to test on YOLO-V4 [48], a state-of-the-art model for real-time object detection. Note that, considering the simplicity of the hardware design, we replace the original activation function Mish with Leaky-ReLU. We adopt Microsoft COCO [49] for the testing dataset, which is widely used for object detection benchmarks. In general, the object detection model is intensely challenging for most quantization methods, as it has fewer prototypical images and needs to cover objects of a broad range.

**Table 2: The object detection benchmark of YOLO-V4 with the baseline full-precision model on COCO dataset based on mean average precision (mAP).**

| Network | Image Size | Method | mAP |
|---|---|---|---|
| YOLO-V4 Leaky | 416x416 | baseline | 62.60% |
| | 416x416 | Ours (without shifting) | 58.30% |
| | 416x416 | **Ours** | **62.00%** |

Using mean average precision (mAP) as an evaluation metric, we compare the full-precision model and our 4-bit quantized model. Thanks to the shifting of the biased activation, Table 2 shows the mAP of our 4-bit quantized YOLO-V4 drops merely 0.6%. This result is very attractive to applications on embedded devices as our model is 4 times smaller while with almost the same model accuracy.

### 5.3 Language Model

We also evaluate the effectiveness of our approach to natural language processing (NLP) tasks by applying our method to recurrent neural networks (RNN) [50][51] for language modeling. RNN is good for computing the probability of a given sequence of words in a sentence and has been widely used in machine translation.

We evaluated a single-layer LSTM (a type of RNN) [52] with 300 hidden units and compared the result with Balanced Quantization [53], which is a uniform quantization method that aims to average the distribution of quantized values in order to maximize utilization per representation. We measure the word-level perplexity (PPL) of these LSTMs before/after our quantization method using the Penn Tree Bank [54] dataset. The perplexity is a measurement of the similarity between the predicted probability distribution and the sample. A lower perplexity score indicates that the probability distribution has a better performance at predicting the sample. Since the baseline model is different in our method and Balanced Quantization, we compared the PPL difference before/after quantization for a fair comparison.

**Table 3: The perplexity (PPL) on LSTM with 300 hidden units, compared with full-precision network and Balanced Quantization. The number in brackets denotes the difference with baseline, the lower the better.**

| | LSTM (300 units) | | | |
|---|---|---|---|---|
| Bits(W/A) | 2/2 | 2/3 | 4/4 | FP/FP (baseline) |
| Balanced | 126 (+19) | 123 (+6) | 114 (+7) | 107 |
| **Ours** | **99.14 (+9.18)** | **92.75 (+2.79)** | **88.41 (-1.55)** | 89.96 |

Table 3 shows that, in all quantization configurations, our method has less degradation (i.e., lower PPL) than Balanced Quantization. Notably, our 4-bit model performs even better than the full-precision LSTM model. In summary, our method provides a comparable prediction capability as the full-precision model but at a much lower hardware cost.